\documentclass[runningheads]{llncs}

\usepackage[T1]{fontenc}
\usepackage{graphicx}
\usepackage{booktabs}
\usepackage{amsmath}
\usepackage{orcidlink}   
\usepackage{placeins}    
\hypersetup{hidelinks}

\graphicspath{{figures/}}

\begin{document}

\title{Reliability analysis for BraTS-GoAT segmentation:\\
       a controlled robustness study of deep-ensemble uncertainty}
\titlerunning{Reliability analysis for BraTS-GoAT segmentation}

\author{Riya Deepak Shet\inst{1}\,\orcidlink{0009-0000-4183-8107} \and
        Chenxi Liang\inst{2} \and
        Le Zhang\inst{3}\,\orcidlink{0000-0002-3848-0017}}
\authorrunning{R. D. Shet et al.}
\institute{Department of Cancer and Genomic Sciences, School of Medical Sciences, College of Medicine and
Health, University of Birmingham, Birmingham, United Kingdom \\ \email{rxs1355@student.bham.ac.uk}
\and
School of Biomedical Engineering, South-Central Minzu University, Wuhan, China
\and
Department of Electronic, Electrical and Systems Engineering, School of Engineering, College of Engineering
and Physical Sciences, University of Birmingham, Birmingham, United Kingdom \\
\email{l.zhang.16@bham.ac.uk}}

\maketitle

\begin{abstract}
Deep networks segment brain tumours accurately in-\hspace{0pt}distribution, but can fail silently when the input
differs from their training data. That risk is central to clinical deployment and is the premise of the
BraTS-GoAT generalizability task. We ask not only how well a model segments, but whether its uncertainty
knows when it is wrong. On BraTS-GoAT (Task~3) we train a 5-fold cross-validated nnU-Net baseline (one
held-out prediction per case) and a 3-seed deep ensemble. Both are evaluated for calibration and error
detection on a per-region relevant mask, aggregated per case. In-distribution the 3-seed ensemble improves
modestly over the already strong single model on the same held-out split, with the clearest gain in
calibration. The separation appears under shift. In a controlled robustness study using graded synthetic
corruptions as a proxy for acquisition shift, the single model's confidence stays flat while its accuracy
and calibration degrade. Inter-member disagreement instead rises steeply, about a quarter to a third above the clean
condition, several times the single model's response. On the official validation leaderboard the 5-fold
ensemble of those folds attains whole-tumour Dice 0.87. The generalization gap is concentrated on the harder
regions, with a characteristic failure of missing small, satellite lesions on unseen cohorts. In the
synthetic study, disagreement among the 3-seed members is a more sensitive case-level indicator of
acquisition shift than single-model confidence. Its per-voxel error localisation weakens as severity grows.
The contribution is a rigorous, honest reliability comparison rather than a claim that any one uncertainty
method dominates.
\keywords{Brain tumour segmentation \and Uncertainty quantification \and Deep ensembles
  \and Calibration \and Robustness \and nnU-Net.}
\end{abstract}

\section{Introduction}
Automated segmentation of gliomas from multiparametric MRI underpins tumour measurement, treatment planning
and response assessment. Modern convolutional pipelines, above all nnU-Net~\cite{isensee2021nnunet}, reach
expert-level overlap on the data distributions they are trained on. The obstacle to clinical use is not
average accuracy but reliability under change. A model deployed on an unfamiliar scanner, protocol or
patient population can degrade while remaining confident. Such a failure is silent, and dangerous precisely
because nothing flags it. The BraTS-GoAT challenge~\cite{goat2026,menze2015brats,baid2021rsna} makes this
concrete by evaluating generalization of tumour sub-region segmentation across heterogeneous tumour types.

We therefore treat the problem as one of \emph{reliability}. Beyond the Dice score we ask whether a
model's own uncertainty lands on its mistakes, and whether it responds when acquisition conditions change.
Two uncertainty signals are available from standard models: the maximum-softmax \emph{confidence} of a
single network, and the \emph{disagreement} between ensemble members. A single nnU-Net trained with Dice and
cross-entropy is known to be overconfident. The question is whether an ensemble's disagreement is a more
trustworthy signal, particularly as acquisition conditions depart from those seen in training.

We make four contributions. The first is a strong nnU-Net baseline for BraTS-GoAT, trained solely on the
challenge data. The second is a paired, per-case comparison of single-model confidence and deep-ensemble
disagreement on calibration and error-detection metrics. The third is a controlled robustness study under
graded synthetic acquisition shift, isolating whether each signal rises as input quality drops. The fourth
is external evaluation on the official
leaderboard.\footnote{Code and evaluation harness: \url{https://github.com/riyashet-hds/brats-goat-reliability}} Prior work
finds that no single uncertainty method dominates~\cite{mehta2022qubrats}. We therefore report an honest
comparison, including where the 3-seed ensemble helps little and where it regresses.

\subsection{Related work}
\textbf{nnU-Net as the standing baseline.} Architectural novelty has repeatedly underperformed careful
configuration in brain tumour segmentation. nnU-Net~\cite{isensee2021nnunet} automates that configuration,
deriving patch size, spacing, normalization and augmentation from a dataset fingerprint. It
has become the standard reference pipeline against which new methods are measured, with strong subsequent
entries refining it rather than departing from it~\cite{isensee2021brats,luu2022extending}. That maturity
makes it a useful fixed substrate. We therefore treat it as a
competitive baseline rather than an axis of contribution, and spend the budget on reliability.

\textbf{Uncertainty benchmarking and its inconclusive verdict.} Uncertainty quantification divides into
epistemic uncertainty, what the model does not know and could learn from more data, and aleatoric
uncertainty, noise inherent in the image. The QU-BraTS benchmark~\cite{mehta2022qubrats} evaluated a field
of methods on brain tumour segmentation. It rewarded uncertainty that suppresses confident errors, and
concluded that no single method dominated. Alongside it, modern
networks are known to be systematically overconfident~\cite{guo2017calibration}, and rankings measured
in-distribution do not survive the shift that matters~\cite{ovadia2019trust}. Those findings set our
design: we compare few signals carefully, on paired cases, and report regressions as well as gains.

\textbf{Deep ensembles against the cheaper approximations.} Independently trained deep
ensembles~\cite{lakshminarayanan2017ensembles} are widely reported as among the strongest uncertainty
estimators, and a common reference point for later methods~\cite{ovadia2019trust}. Their cost is $K$ full
training runs. Two cheaper approximations are standard. Monte-Carlo dropout (MC-dropout) samples
predictions with dropout left active at inference~\cite{gal2016dropout}. Test-time augmentation (TTA)
treats the spread over acquisition-model transformations as uncertainty~\cite{wang2019tta}. Both
obtain a distribution from a single trained network, so both are weaker proxies for genuine model
diversity. We therefore scope the study to the strongest estimator
and the baseline it must beat, naming MC-dropout and TTA as the next comparators.

A parallel line of work models the same degradations in order to undo them. Examples include imputation of
missing slices~\cite{zhang2019missing}, joint restoration of blur, bias field and
noise~\cite{zhang2022learning}, and diffusion-based tumour inpainting~\cite{tao2025diffkan}. We use
degradation as a controlled probe rather than a target for restoration.

\section{Materials and Methods}

\subsection{Data and task}
We use BraTS-GoAT Task~3, the generalizability track. It comprises 1351 labelled training cases, each with
four co-registered MRI sequences (native and contrast-enhanced T1, T2-weighted, T2-FLAIR) at
$240\times240\times155$ and $1\,\mathrm{mm}^3$
resolution~\cite{goat2026,bakas2017advancing,bratsgli2024posttreatment}. The challenge data and its
evaluation are distributed through the MedPerf federated benchmarking
platform~\cite{karargyris2023medperf}. Evaluation uses the three nested BraTS regions: whole tumour (WT),
tumour core (TC) and enhancing tumour (ET). About 2.4\% of cases (33) contain no enhancing tumour, where
over-segmentation is penalised heavily and which are salient for error detection. Per the GoAT rules, the
model is trained only on the provided data, from scratch, with no external data or pretrained weights.

\subsection{Segmentation baseline}
\label{sec:baseline}
The baseline is nnU-Net in its ResEnc-L 3d\_fullres configuration~\cite{isensee2021nnunet}, trained from
scratch under 5-fold cross-validation. Pooling the five held-out folds yields one prediction per case from a
model that did not train on it. That gives the cross-validated single model over all 1351 cases. The same
five networks are also used a second way: for the challenge submission all five are averaged at inference,
giving the 5-fold ensemble. The two configurations are not interchangeable.

We use nnU-Net v2.8.0 with the stock trainer and residual-encoder-L plans. The network is a six-stage
residual-encoder U-Net, 32 to 320 features, patch $160\times192\times160$, batch 3, trained 1000 epochs per
fold (about 29\,h on an A100 of our BlueBEAR cluster). Every other setting is the library default:
stochastic gradient descent with Nesterov momentum 0.99, polynomial decay from $10^{-2}$, and a Dice and
cross-entropy loss with deep supervision. Preprocessing is z-score normalisation inside the brain mask at the native $1\,\mathrm{mm}$
spacing, so no resampling occurs. Training uses the four tissue labels, with nested regions derived at
evaluation. No post-processing is applied, so all numbers are raw. One inference setting
differs between runs. nnU-Net's end-of-training validation enables test-time mirroring by default, so the
internal cross-validated numbers include it. Both the leaderboard predictions and the submitted container
run with it disabled.

\subsection{Deep ensemble}
The uncertainty method is a 3-seed deep ensemble: three nnU-Nets trained on a single fixed split with
different random seeds~\cite{lakshminarayanan2017ensembles}. The seeds are nnU-Net's default and two
explicit alternatives. Because the members share one training split, their held-out cases are unseen by every member,
so their disagreement is a clean, label-referenced uncertainty. The 5-fold ensemble lacks this
property, since its members' training sets overlap. The paired baseline is the fold-0 network of
Section~\ref{sec:baseline} on that split's 271 held-out cases, the split-0 single model.
We report two signals: maximum-softmax confidence (available to both) and inter-member disagreement
(ensemble only). We use three members due to compute reasons: at about 29\,h per training run, a
$K$-member ensemble costs $K$ full runs.

Three model configurations arise from two training regimes. The cross-\hspace{0pt}validated single model of
Section~\ref{sec:baseline} provides the internal baseline: each case is predicted by the fold that held it
out. The 5-fold challenge ensemble averages those same fold models at inference, is packaged in the
container, and produces the leaderboard results of Section~\ref{sec:external} only. Its members' training
sets overlap, so it is not used for label-referenced uncertainty. The 3-seed deep ensemble, paired with a
single model on the same held-out split, carries the reliability and robustness comparisons. The
leaderboard therefore anchors external segmentation performance, not the uncertainty analysis.

\textbf{Uncertainty signals.} Reliability quantities use a binary per-region probability from the four-class
softmax: $P(\mathrm{WT}) = 1 - P(\mathrm{bg})$, $P(\mathrm{TC}) = P(\mathrm{NCR}) + P(\mathrm{ET})$, and
$P(\mathrm{ET})$ directly. At a voxel $v$, \emph{confidence} is that of the decided class,
$c(v) = \max\bigl(P(v),\,1 - P(v)\bigr)$, with uncertainty $1 - c(v)$. \emph{Disagreement} is the variance
$D(v) = \frac{1}{K}\sum_{m}\bigl(P_m(v) - \bar{P}(v)\bigr)^2$ of the $K = 3$ members about their mean
$\bar{P}(v) = \frac{1}{K}\sum_{m} P_m(v)$, which is also the 3-seed ensemble's predictive probability. Both are
evaluated only on the relevant mask defined below. Being a variance of a probability, $D$ is inherently
small, so it is reported as a relative change.

\subsection{Reliability metrics}
Segmentation quality uses Dice, the 95th-percentile Hausdorff distance (HD95, mm) and normalized surface
Dice (NSD)~\cite{maierhein2024metrics}. Calibration uses the expected calibration error
(ECE)~\cite{guo2017calibration}. Error detection uses the AUROC of uncertainty against per-voxel error and
the area under the risk--coverage curve (AURC)~\cite{geifman2017selective}. Following
QU-BraTS~\cite{mehta2022qubrats}, calibration and error-detection metrics are computed on the per-region
relevant mask (the union of predicted and reference region, lightly dilated) and aggregated per case then
averaged. The relevant mask removes the class-imbalance dilution that makes all-brain ECE uninformative.
Per-case aggregation avoids a Simpson's-paradox inversion of AUROC across cases.

Concretely, the relevant mask dilates the union of the reference region and the thresholded prediction
$P(r) \ge 0.5$ by two iterations of a six-connected element. ECE uses 15 equal-width bins. At most
20\,000 mask voxels per case and region are sampled, and AUROC is dropped for cases with no errors or
only errors. Tables report per-case ECE, whereas Fig.~\ref{fig:reliability} and the values quoted from
it are pooled, so the two differ slightly by construction.

\subsection{Controlled robustness study}
GoAT anonymises the per-case cohort, so we cannot partition the labelled data by tumour type. We instead run
a controlled robustness study. The validation images are degraded by four MRI-realistic corruptions
(Gaussian noise, bias field, blur, gamma) at graded severity, with the reference segmentation held fixed,
and inference is re-run. Because the corruption and its strength are controlled exactly, this isolates
whether the model and its uncertainty respond as input quality drops. It is a controlled proxy for
acquisition shift, not a substitute for validation on genuinely unseen cohorts, and we treat it as such.

Additive Gaussian noise uses $\sigma$ of 0.05, 0.10 and 0.20 of each channel's brain-intensity standard
deviation. The multiplicative bias field has a peak fractional swing of 0.3, 0.5 and 0.8. Blur is an
isotropic Gaussian of $\sigma$ 0.5, 0.75 and 1.0 voxels. Gamma remaps brain intensities rescaled to
$[0,1]$ with exponents 0.8 and 1.25, a remap rather than an ordinal ladder, so its settings are categorical.
With a shared clean condition this gives twelve in total. Corruptions are applied to all four modalities,
inside the brain mask, before nnU-Net's normalisation, so the model sees genuinely degraded input rather
than a rescaling it would undo. For each corruption and severity we draw one fixed random realisation and
apply it to every case, so cases differ in anatomy but not in the perturbation received. Severity was capped
by inspecting previews, and the study runs on the 271 held-out cases of the fixed split.

\section{Results}

\subsection{Segmentation baseline}
\begin{figure}[t]
\centering
\includegraphics[width=0.78\linewidth]{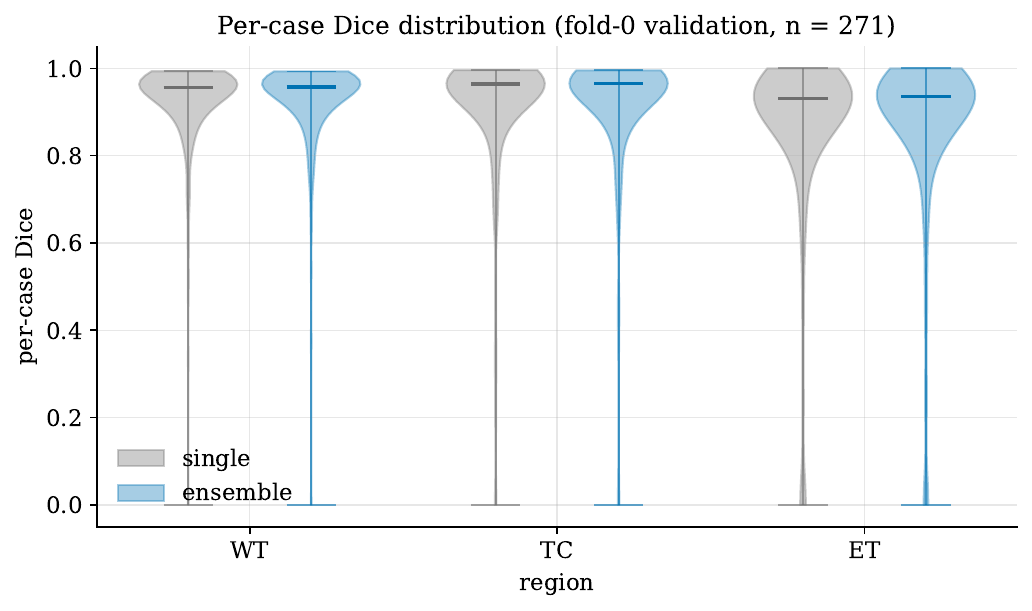}
\caption{Per-case Dice distributions on the internal validation split ($n=271$), split-0 single model vs
3-seed deep ensemble, by region. The mass concentrates near 1.0 with thin tails reaching 0, so the typical
case is excellent and the mean is set by a hard-case tail.}
\label{fig:violin}
\end{figure}

The cross-validated single model is a strong baseline over all 1351 cases (Table~\ref{tab:baseline}). Per-case
medians sit well above the means, and median HD95 is roughly 1\,mm against means of 6--11\,mm. The typical
case is therefore excellent, and the means reflect a hard-case tail (Fig.~\ref{fig:violin}). Mean ET HD95 is
10.73\,mm over all cases but 5.62\,mm on ET-present cases. The 33 no-ET cases, where the model hallucinates a
small enhancing region, drive most of that boundary error.

\begin{table}[t]
\centering
\caption{Cross-validated single-model segmentation quality (nnU-Net ResEnc-L, 5-fold, $n=1351$), distinct
from the 5-fold challenge ensemble used for Section~\ref{sec:external}. Means with per-case medians in
parentheses. Dice/NSD higher is better;
HD95 (mm) lower is better. The final row restricts ET to the 1318 ET-present cases.}
\label{tab:baseline}
\begin{tabular}{lccc}
\toprule
Region & Dice & HD95 (mm) & NSD \\
\midrule
WT & 0.925 (0.958) & 6.45 (1.73) & 0.858 (0.913) \\
TC & 0.909 (0.964) & 6.29 (1.17) & 0.865 (0.950) \\
ET & 0.869 (0.936) & 10.73 (1.00) & 0.891 (0.962) \\
\midrule
ET (present only) & 0.880 (0.936) & 5.62 (1.00) & 0.903 (0.962) \\
\bottomrule
\end{tabular}
\end{table}

\subsection{Calibration and error detection}
On the relevant mask, over all 1351 cases, the cross-validated single model is overconfident. The reliability
curves lie below the diagonal in every region (Fig.~\ref{fig:reliability}), with per-case ECE
0.070 / 0.076 / 0.080 (WT/TC/ET). Its confidence is a weak-but-genuine error detector, with AUROC
0.875 / 0.872 / 0.857 and AURC around 0.03--0.04, exceeding chance in 98--99\% of cases. Overconfidence with
usable-but-imperfect ranking is the gap an ensemble should close.

\subsection{3-seed ensemble versus split-0 single model}
\begin{figure}[b]
\centering
\includegraphics[width=\linewidth]{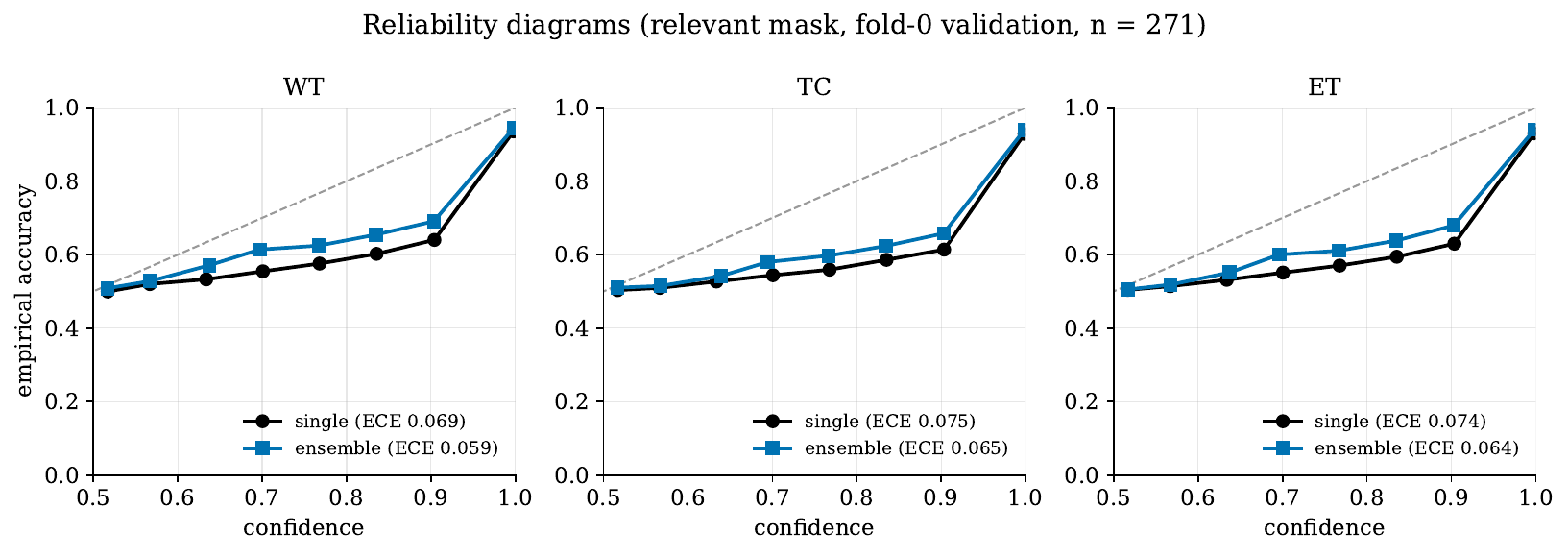}
\caption{Reliability diagrams on the per-region relevant mask ($n=271$). Both curves lie below the
perfect-calibration diagonal, so both models are overconfident, as expected for nnU-Net. The 3-seed ensemble
(blue) sits consistently closer to the diagonal, with lower pooled ECE in every region (legend); this
calibration gain is its clearest in-distribution advantage.}
\label{fig:reliability}
\end{figure}

On the 271 held-out cases of the fixed split the 3-seed ensemble is significantly better on almost every
metric, but the effect sizes are small (Table~\ref{tab:paired}). The clearest gain is calibration. Pooled
ECE falls from 0.069 / 0.075 / 0.074 for the split-0 single model to 0.059 / 0.065 / 0.064. The 3-seed
curve sits closer to the diagonal in Fig.~\ref{fig:reliability}. The paired Wilcoxon test is run on the per-case deltas of
Table~\ref{tab:paired} (all $p<0.001$). Dice improves by 0.003--0.007 and AUROC by 0.004--0.009. This
paired test compares the two models on their \emph{confidence} signal, the one both possess. Disagreement
has no single-model counterpart, so it enters as the descriptive shift-response result in
Section~\ref{sec:robust}. The boundary effect is region-specific rather than a blanket cost. The 3-seed
ensemble significantly improves ET HD95 (13.59 to 12.14\,mm, $p=0.035$) but worsens TC HD95 (3.14 to
4.27\,mm, $p=0.038$), with WT unchanged. In-distribution, then, the 3-seed ensemble is a modest,
calibration-focused improvement over the split-0 single model.

The pattern differs by region in a way worth naming. Calibration improves almost identically everywhere
($\Delta$ECE $-0.010$ to $-0.011$), so averaging helps globally. Boundary distance does not: the ensemble
helps ET, whose errors are large, and slightly hurts TC, whose boundary is already tight at a median HD95
near 1\,mm. Error detection improves most on TC ($+0.009$) and least on WT ($+0.004$). ET is the hardest
region throughout, with the lowest Dice and worst calibration. Its aggregate boundary error is driven
almost entirely by the no-ET cases. Mean ET HD95 falls from 10.73\,mm to 5.62\,mm once they are excluded.

\begin{table}[t]
\centering
\caption{Paired comparison, 3-seed deep ensemble minus the split-0 single model, on the 271 held-out cases
($\Delta = \text{ensemble} - \text{single}$; Wilcoxon signed-rank; $^{*}$ denotes $p<0.05$). Negative
$\Delta$ is an improvement for HD95/ECE/AURC.}
\label{tab:paired}
\begin{tabular}{lccc}
\toprule
Metric ($\Delta$) & WT & TC & ET \\
\midrule
Dice          & $+0.003^{*}$ & $+0.004^{*}$ & $+0.007^{*}$ \\
NSD           & $+0.004^{*}$ & $+0.006^{*}$ & $+0.008^{*}$ \\
HD95 (mm)     & $-0.01$ (n.s.) & $+1.14^{*}$ & $-1.46^{*}$ \\
ECE           & $-0.010^{*}$ & $-0.011^{*}$ & $-0.010^{*}$ \\
AUROC (error, confidence) & $+0.004^{*}$ & $+0.009^{*}$ & $+0.007^{*}$ \\
AURC          & $-0.002^{*}$ & $-0.004^{*}$ & $-0.003^{*}$ \\
\bottomrule
\end{tabular}
\end{table}

\subsection{Robustness under synthetic acquisition shift}
\label{sec:robust}
\begin{figure}[t]
\centering
\includegraphics[width=0.92\linewidth]{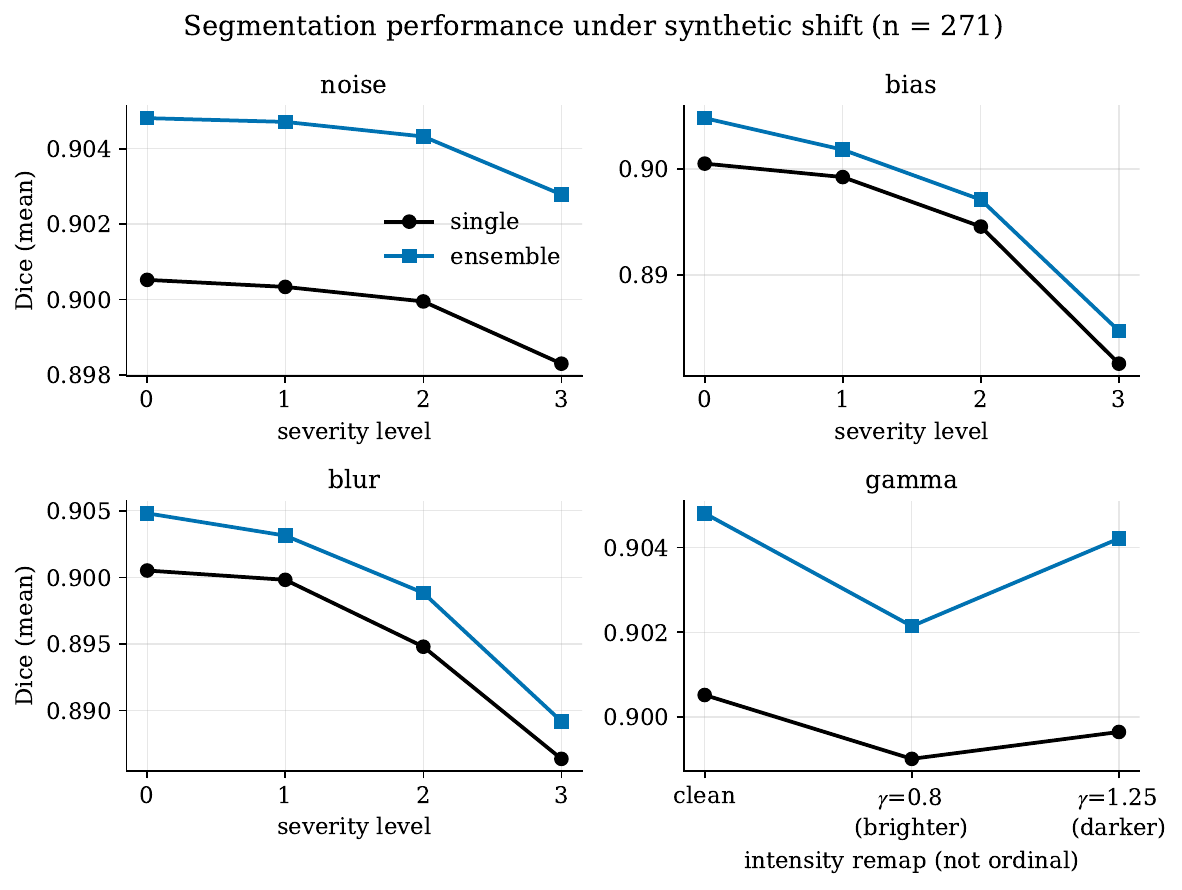}
\caption{Segmentation performance under the controlled robustness study ($n=271$), split-0 single model vs
3-seed deep ensemble, one panel per corruption. Dice is averaged over WT/TC/ET. Note the narrow $y$-axis:
the drop is small in absolute terms even at severe bias and blur, and noise and gamma barely move. Gamma
is an intensity remap, not an ordinal severity ladder, so its $x$-axis is categorical. These are
synthetic, graded corruptions, a controlled proxy for acquisition shift rather than real cohort shift.}
\label{fig:stressdice}
\end{figure}

\begin{figure}[t]
\centering
\includegraphics[width=\linewidth]{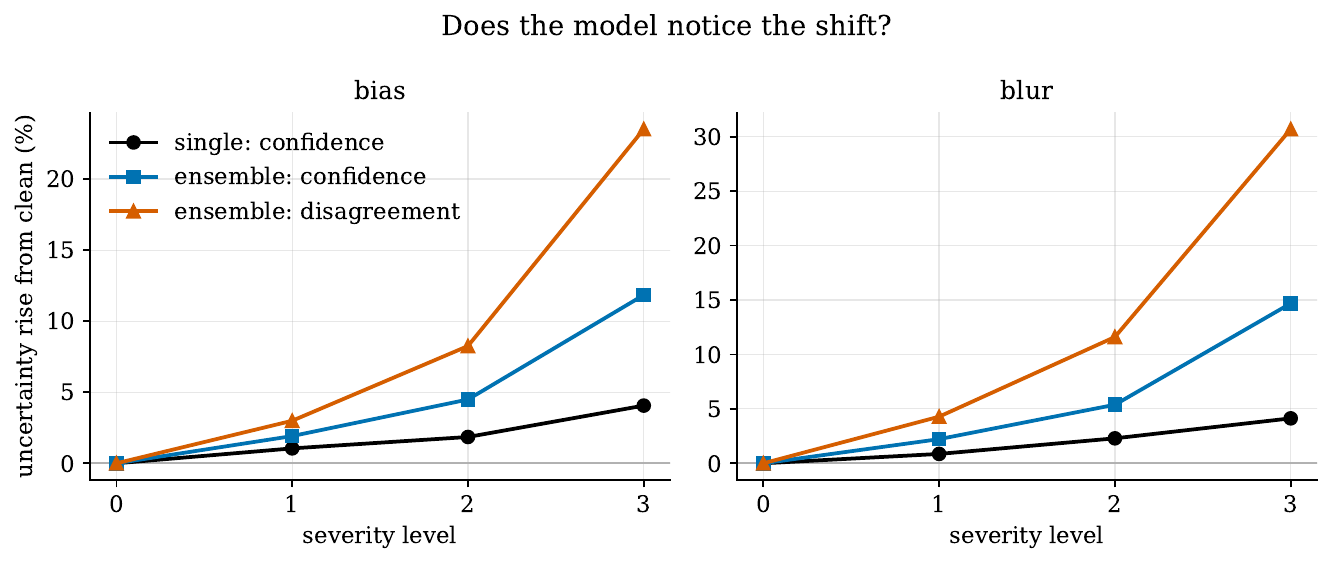}
\caption{Percentage rise of each uncertainty signal from the clean condition, for the two corruptions that
degrade Dice (bias, blur). Single-model confidence (black) barely moves; 3-seed confidence (blue) rises
more; 3-seed disagreement (orange) rises steeply, to about $+23\%$ under severe bias and $+31\%$ under
severe blur. The 3-seed ensemble's disagreement responds to degradation that single-model confidence
absorbs unchanged.}
\label{fig:stressdetect}
\end{figure}

The robustness study separates the split-0 single model from the 3-seed ensemble
(Fig.~\ref{fig:stressdice},~\ref{fig:stressdetect}).
Segmentation quality degrades gracefully. Bias field and blur bite most, with the split-0 single model's
region-averaged Dice falling from
0.900 to 0.882 under severe bias, while noise and gamma barely register. The informative result is the
uncertainty. The single model's confidence is nearly flat as quality drops. Its mean uncertainty rises only
about 5\% under severe bias, while its ECE on that split climbs about 25\% (0.077 to 0.096).
It becomes worse and more
miscalibrated while remaining equally confident, which is the silent-failure signature. Inter-member
disagreement, by
contrast, rises steeply, about $+23\%$ under bias and $+31\%$ under blur, several times the single model's
response. The sensitivity order on identical data is disagreement $\gg$ 3-seed confidence $>$ single
confidence. The error-detection AUROC dips slightly, from about 0.87 to
0.86. Disagreement is therefore the more sensitive case-level indicator
of corruption-induced shift, even though its per-voxel error localisation weakens as severity grows.

\subsection{External evaluation on the challenge leaderboard}
\label{sec:external}
We submitted predictions for all 451 cases of the official BraTS-GoAT validation set (the returned per-case
score file covered 450 of them). The 5-fold ensemble attains mean per-case Dice 0.875 (WT), 0.814 (TC) and
0.782 (ET) over 450, 449 and 435 scored cases. The differing denominators reflect regions the official
pipeline leaves unscored. Bootstrap 95\% intervals are $[0.856,0.892]$, $[0.789,0.838]$, $[0.753,0.809]$.
Mean NSD is 0.48 / 0.50 / 0.55 and mean lesion-wise $F_1$ 0.69 / 0.74 / 0.67. Mean HD95 is
17.1 / 25.0 / 40.0\,mm against medians of 2.2 / 2.0 / 1.4\,mm. The HD95 mean is itself unstable
under resampling where the median is not. Mean whole-tumour HD95 spans 11.5 to 23.6\,mm under the same
resampling, its median only 2.24 to 2.45\,mm. That is the catastrophic tail seen from outside. The Dice gap
against the internal numbers, 0.050 / 0.095 / 0.087, is two to four times the interval half-width and so
is not a sampling artefact. The two sides use different inference
configurations: the internal estimate is the cross-validated single model, the leaderboard the 5-fold
challenge ensemble. Ensembling would raise the leaderboard score, disabling test-time mirroring would lower
it. The gap is therefore an external anchor rather than a protocol-matched estimate, broadly consistent
with the synthetic robustness trend but not directly comparable. Lesion-wise metrics are driven by
under-detection of small satellite lesions rather than uniform decline. A case-level signal should
ideally flag that, though the challenge outputs did not let us test it.

\FloatBarrier
\section{Discussion}
Two findings stand out. First, in-distribution a single nnU-Net is already a strong
reliability baseline. The 3-seed deep ensemble adds small, mostly calibration-focused gains rather than a
decisive advantage, echoing QU-BraTS's conclusion that no single method
dominates~\cite{mehta2022qubrats}. The exception is boundary distance: ET HD95 improves, but TC HD95
slightly worsens. Because HD95 is set by the worst boundary tail rather than by average overlap, small
changes in outlying points move it without implying a uniform improvement. A possible mechanism, which we
did not test, is that averaging pulls in far outliers while smoothing a boundary that was already
correct. Second, the 3-seed ensemble's value is clearest under corruption-induced shift. Its disagreement
rises when the single model's confidence does not, which is why it is the more shift-sensitive signal.

A case-level error analysis explains the gap between the strong medians and the weaker means. Median Dice is
0.94 / 0.94 / 0.90. Removing the 6--12\% of cases with HD95 above 50\,mm drops mean HD95 to
3.5 / 4.5 / 2.8\,mm. Two
failure modes account for it. The first is under-detection of multifocal disease, in 16--19\% of cases. The
second is that errors are dominated by omission rather than hallucination. Mean false positives per
case are 0.3--1.1 against false negatives of 1.6--2.5, so the model misses tumour quietly. This is the
silent-failure profile the analysis targets.

Several limitations bound these claims. The controlled robustness study uses synthetic corruptions as a
stand-in for real cohort shift. The leaderboard result corroborates those trends on a genuinely unseen
cohort rather than validating them. The 3-seed ensemble comparison is on one held-out split, and its gains,
while statistically significant, are small in absolute terms. Disagreement is small in magnitude, so its
value is the relative rise under corruption-induced shift rather than an absolute threshold. Corruption
severity was capped by our own judgement, and each condition uses one fixed realisation. Reliability was
reported for one fold, so we also computed it per fold. The five held-out sets hold 271, 270, 270, 270 and
270 cases and sum to 1351. Across them ECE is 0.0697 $\pm$ 0.0028 (WT), 0.0761 $\pm$ 0.0051 (TC) and
0.0798 $\pm$ 0.0043 (ET), as mean $\pm$ SD. Detection AUROC is 0.8748 $\pm$ 0.0075, 0.8716 $\pm$ 0.0056
and 0.8571 $\pm$ 0.0050. Calibration and detection are therefore stable across all five folds. The
ordering between regions is not: calibration
WT $<$ TC $<$ ET holds in four folds and detection WT $>$ TC $>$ ET in three. Only ET having the lowest
Dice and detection AUROC survives all five. These are single-model results, so the ensemble-versus-single difference is still a
single-split measurement. Future work, scoped out to keep the comparison focused, includes
MC-dropout and TTA as additional estimators, and validation on real external cohorts. Semi-supervised
training is a natural extension, but GoAT permits no external data, and pseudo-labels inject
error-correlated noise whose effect on calibration is itself an open question.

\begin{credits}
\subsubsection{\ackname}
Computation used the University of Birmingham BlueBEAR HPC service. Data came from the
BraTS-GoAT challenge (Synapse ID syn74274097).

\subsubsection{\discintname}
The authors have no competing interests to declare.
\end{credits}

\bibliographystyle{splncs04}
\bibliography{refs}

\end{document}